\documentclass[11pt,a4paper]{article}
\usepackage[utf8]{inputenc}
\usepackage[T1]{fontenc}

\usepackage[hyperref]{acl2020}
\usepackage{times}
\usepackage{latexsym}

\usepackage{microtype}
\usepackage{amsmath}
\usepackage{xcolor}
\usepackage{graphicx}
\usepackage{enumitem}
\usepackage{booktabs}
\usepackage{multirow}
\usepackage{array}
\usepackage{float}
\usepackage{arydshln}
\usepackage{adjustbox}
\usepackage{multicol}
\usepackage{mathabx}
\usepackage{authoraftertitle}
\usepackage{pbox}
\usepackage{arydshln}
\usepackage[absolute]{textpos}

\newcommand{\redund}[1]{{\color{black!30!red!100}\underline{#1}}}
\newcommand{\greenund}[1]{{\color{black!40!green!100}\underline{#1}}}

\usepackage[normalem]{ulem}

\def\ODdel#1{\bgroup\markoverwith{\textcolor{cyan!80!yellow!80!black}{\rule[0.5ex]{2pt}{1pt}}}\ULon{#1}}

\aclfinalcopy 

\title{Data-to-Text Generation with Iterative Text Editing}

\author{Zdeněk Kasner \and Ondřej Dušek \\
  Charles University, Faculty of Mathematics and Physics\\
  Institute of Formal and Applied Linguistics \\
  Prague, Czech Republic \\
  \texttt{\{kasner,odusek\}@ufal.mff.cuni.cz}
}

\date{}

\begin{document}

\maketitle

\begin{textblock*}{\textwidth}(2.5cm,1cm)
\noindent
In \emph{Proceedings of the 13th International Conference on Natural Language Generation (INLG)}, Online, December 2020.
\end{textblock*}

\begin{abstract}
We present a novel approach to data-to-text generation based on iterative text editing. Our approach maximizes the completeness and semantic accuracy of the output text while leveraging the abilities of recent pre-trained models for text editing (\textsc{LaserTagger}) and language modeling (GPT-2) to improve the text fluency. To this end, we first transform data items to text using trivial templates, and then we iteratively improve the resulting text by a neural model trained for the \textit{sentence fusion} task. The output of the model is filtered by a simple heuristic and reranked with an off-the-shelf pre-trained language model. We evaluate our approach on two major data-to-text datasets (WebNLG, Cleaned E2E) and analyze its caveats and benefits. Furthermore, we show that our formulation of data-to-text generation opens up the possibility for zero-shot domain adaptation using a general-domain dataset for sentence fusion. \footnote{The code for the experiments is available at \url{https://github.com/kasnerz/d2t_iterative_editing}}
\end{abstract}

\section{Introduction}
Data-to-text (D2T) generation is the task of transforming structured data into a natural language text which represents it \citep{reiter2000building,gatt2018survey}. 
The output text can be generated in several steps following a pipeline, or in an end-to-end (E2E) fashion. Neural-based E2E architectures recently gained attention due to their potential to reduce the human input needed for building D2T systems. A disadvantage of E2E architectures is the lack of intermediate steps, which makes it hard to control the semantic fidelity of the output \citep{moryossef2019astep,ferreira2019neural}.

We focus on a D2T setup where the input data is a set of RDF triples in the form of \textit{(subject, predicate, object)} and the output text represents \textit{all} and \textit{only} facts in the data. This setup can be used by all D2T applications where the data describe relationships between entities  \cite[e.g.][]{gardent2017webnlg,budzianowski_multiwoz_2018}.\footnote{The setup can be preceded by the \textit{content selection} for selecting the relevant subset of data \cite[cf.][]{wiseman_challenges_2017}.} In order to combine the benefits of pipeline and E2E architectures, we propose to use the neural models with a limited scope. We take advantage of three facts: (1) each triple can be lexicalized using a trivial template, (2) stacking the lexicalizations one after another tends to produce an unnatural sounding but semantically accurate output, and (3) the neural model can be used for combining the lexicalizations to improve the output fluency.

In traditional pipeline-based NLG systems \citep{reiter2000building}, combining the lexicalizations is a non-trivial multi-stage process. Text structuring and sentence aggregation are first used to determine the order of facts and their assignment to sentences, followed by referring expression generation and linguistic realization. We argue that with a neural model, combining the lexicalizations can be simplified as several iterations of \textit{sentence fusion}---a task of combining sentences into a coherent text \citep{barzilay2005sentence}.

Our contributions are the following: 
\begin{enumerate}[label={\arabic*)},nosep]
\item We show how to reframe D2T generation as iterative text editing, which makes it independent of dataset-specific input data format and allows to control the output over a series of intermediate steps.
\item We perform initial experiments using our approach on two major D2T datasets (WebNLG and Cleaned E2E) and include a quantitative and qualitative analysis of the results.
\item We perform zero-shot domain adaptation experiments and show that our approach exhibits a domain-independent behavior.

\end{enumerate}

\begin{figure*}[t]
    \centering 
    \includegraphics[trim={0.1cm 0.3cm 0.2cm 0},clip,width=\textwidth]{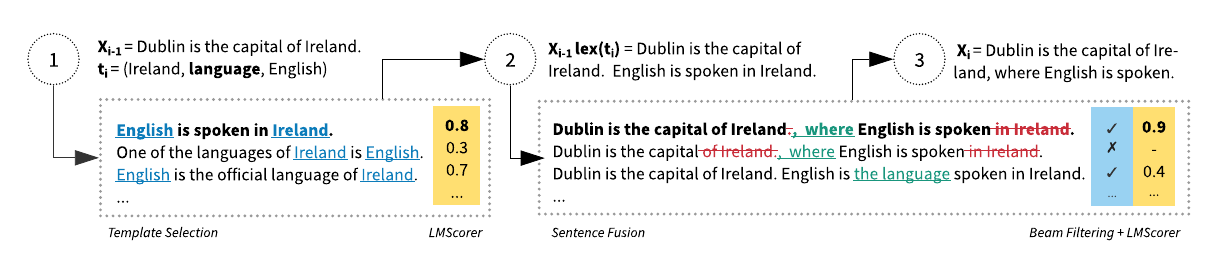}
    \caption{An example of a single iteration of our algorithm for D2T generation. In Step 1, the template for the triple is selected and filled. In Step 2, the sentence is fused with the template. In Step 3, the result for the next iteration is selected from the beam by filtering and language model scoring.}\label{fig:alg}
\end{figure*}

\section{Background}

Improving the accuracy of neural D2T approaches has attracted a lot of research interest lately.
Similarly to us, other systems use a generate-then-rerank approach \citep{dusek_sequence--sequence_2016,juraska_deep_2018} or a classifier to filter incorrect output \citep{harkous2020have}. 
\citet{moryossef2019bimproving,moryossef2019astep} 
split the D2T process into a symbolic text-planning stage and a neural generation stage. Other works improve the robustness of the neural model \citep{tian2019sticking,kedzie_good_2019} or employ a natural language understanding model \citep{nie_simple_2019} to improve the faithfulness of the output. Recently, \citet{chen-etal-2020-shot} finetuned GPT-2 \citep{radford2019language} for a few-shot domain adaptation.

Several models were recently applied to generic text editing tasks. \textsc{LaserTagger} \citep{malmi2019encode}, which we use in our approach, is a sequence tagging model based on the Transformer \citep{vaswani2017attention} architecture with the BERT \citep{devlin-etal-2019-bert} pre-trained language model as the encoder. Other recent text-editing models without a pre-trained backbone include EditNTS \citep{dong-etal-2019-editnts} and Levenshtein Transformer \citep{gu2019levenshtein}.

Concurrently with our work, \citet{kale2020few} explored using templates for dialogue response generation. 
They use the sequence-to-sequence T5 model \citep{raffel2019exploring} to generate the output text from scratch instead of iteratively editing the intermediate outputs, which leaves less control over the model.

\section{Our Approach}
We start from single-triple templates and iteratively fuse them into the resulting text while filtering and reranking the results. We first detail the main components of our system and then give an overall description of the decoding algorithm.

\subsection{Template Extraction}
We collect a set of templates for each predicate. The templates can be either handcrafted, or automatically extracted from the lexicalizations of the single-triple examples in the training data. For unseen predicates, we add a single fallback template: \textit{The <predicate> of <subject> is <object>.}

\subsection{Sentence Fusion}
We train an in-domain sentence fusion model.
We select pairs $(X, X')$ of examples from the training data consisting of $(n, n+1)$ triples and having $n$ triples in common. This leaves us with an extra triple $t$ present only in $X'$. To construct the training data, we use the concatenated sequence $X~\mathrm{lex}(t)$ as a source and the sequence $X'$ as a target, where $\mathrm{lex}(t)$ denotes lexicalizing the triple $t$ using an appropriate template. 
As a result, the model learns to integrate $X$ and $t$ into a single coherent expression.

We base our sentence fusion model on \textsc{LaserTagger} \citep{malmi2019encode}. \textsc{LaserTagger}
is a sequence generation model which generates outputs by tagging inputs with edit operations: \texttt{KEEP} a token, \texttt{DELETE} a token, and \texttt{ADD} a phrase before the token. In tasks where the output highly overlaps with the input, such as sentence fusion, \textsc{LaserTagger} is able to achieve performance comparable to state-of-the-art models with faster inference times and less training data.

An important feature of \textsc{LaserTagger} is the limited size of its vocabulary, which consists of $l$ most frequent (possibly multi-token) phrases used to transform inputs to outputs in the training data. After the vocabulary is precomputed, all infeasible examples in the training data are filtered out. At the cost of limiting the number of training examples, this filtering makes the training data cleaner by removing outliers. The limited vocabulary also makes the model less prone to common neural model errors such as hallucination, which allows us to control the semantic accuracy to a great extent using only simple heuristics and language model rescoring.

\subsection{LM Scoring}
We use an additional component for calculating an indirect measure of the text fluency. We refer to the component as the \textsc{LMScorer}. In our case,  \textsc{LMScorer} is a pre-trained GPT-2 language model \citep{radford2019language} from the Transformers repository\footnote{\url{https://github.com/huggingface/transformers}} \citep{Wolf2019HuggingFacesTS} wrapped in the \textit{lm-scorer}\footnote{\url{https://github.com/simonepri/lm-scorer}} package. We use \textsc{LMScorer} to compute the score of the input text $X$ composed of tokens $x_1\ldots x_n$ as a geometric mean of the token conditional probability:
\begin{align*}
    \operatorname{score}(X) = \Bigg( \prod_{i=1}^{n}{P(x_i|x_1 \ldots x_{i-1})} \Bigg)^{\frac{1}{n}}.
\end{align*}

\begin{table}[t]
    \centering\small
    \begin{tabular}{ll}
        \multicolumn{2}{c}{\textbf{WebNLG}}  \\[0.1cm]
     \textit{foundedBy} &         \pbox{10cm}{<obj> was the founder of <subj>.\\<subj> was founded by <obj>.} \\[0.2cm] \midrule  \\[-0.3cm]
        \multicolumn{2}{c}{\textbf{E2E (extracted)}} \\[0.1cm]
        \textit{area+food} &   \pbox{5.5cm}{<subj> offers <obj2> cuisine in the <obj1>. \\ <subj> in <obj1> serves <obj2> food.} \\[0.2cm] \midrule \\[-0.3cm]
        \multicolumn{2}{c}{\textbf{E2E (custom)}} \\[0.1cm]
        \textit{near} &   \pbox{5cm}{<subj> is located near <obj>. \\ <obj> is close to <subj>.} 
    \end{tabular}
    \caption{Examples of templates we used in our experiments. The templates for the single predicates in the WebNLG dataset and the pairs of predicates in the E2E dataset are extracted automatically from the training data; the templates for the single predicates in E2E are created manually.}
    \label{tab:templates_ex}
\end{table}

\subsection{Decoding Algorithm}
\label{sec:alg}
The input of the algorithm (Figure~\ref{fig:alg}) is a set of $n$ ordered triples. First, we lexicalize the triple $t_0$ to get the base text $X_0$. We choose the lexicalization for the triple as the filled template with the best score from \textsc{LMScorer}. This promotes templates which sound more natural for particular values. 
In the following steps $i=1\ldots n$, we lexicalize the triple $t_i$ and append it after $X_{i-1}$. We feed the joined text into the sentence fusion model and produce a beam with fusion hypotheses. We use a simple heuristic (string matching) to filter out hypotheses in the beam missing any entity from the input data. Finally, we rescore the remaining hypotheses in the beam with \textsc{LMScorer} and let the hypothesis with the best score be the base text $X_{i}$. In case there are no sentences left in the beam after the filtering step, we let $X_{i}$ be the text in which the lexicalized $t_i$ is appended after $X_{i-1}$ without fusion (preferring accuracy to fluency). The output of the algorithm is the base text $X_n$ from the final step.


\begin{table*}[t]
    \centering\small
    \begin{tabular}{lccccc<{\hspace{2mm}}c>{\hspace{2mm}}ccccc} \toprule
     & \multicolumn{5}{c}{\bf WebNLG} & & \multicolumn{5}{c}{\bf Cleaned E2E} \\
    \cmidrule{2-6} \cmidrule{8-12}
     & {\it BLEU} & {\it NIST} & \hspace{-1mm}{\it METEOR}\hspace{-1mm} & \hspace{-1mm}{\it ROUGE$_L$}\hspace{-1mm} & \hspace{-1mm}{\it CIDEr}\hspace{-1mm} & & {\it BLEU} & {\it NIST} & \hspace{-1mm}{\it METEOR}\hspace{-1mm} & \hspace{-1mm}{\it ROUGE$_L$}\hspace{-1mm} & \hspace{-1mm}{\it CIDEr}\hspace{-1mm} \\
    {\bf baseline} & 0.277 & 6.328 & 0.379 & 0.524 & 1.614 & & 0.207 & 3.679 & 0.334 & 0.401 & 0.365 \\
    {\bf zero-shot } & 0.288 & 6.677 & 0.385 & 0.530 & 1.751 & & 0.220 & 3.941 & 0.340 & 0.408 & 0.473 \\
    {\bf w/fusion } & 0.353 & 7.923 & 0.386 & 0.555 & 2.515 & & 0.252 & 4.460 & 0.338 & 0.436 & 0.944 \\
    {\bf SFC } & 0.524 & - & 0.424 & 0.660 & 3.700 & & 0.436 & - & 0.390 & 0.575 & 2.000 \\
    {\bf T5 } & 0.571 & - & 0.440 & - & - & & -  & - & - & - & - \\ \bottomrule
    \end{tabular}
    \caption{Results of automatic metrics on the WebNLG and Cleaned E2E test sets. The comparison is made with the results from the papers on the Semantic Fidelity Classifier (SFC; \citealp{harkous2020have}) and the finetuned T5 model (T5; \citealp{kale2020text}).}
    \label{tab:results}
    \end{table*}

\section{Experiments}

\subsection{Datasets} The WebNLG dataset \citep{gardent2017webnlg} consists of sets of DBPedia RDF triples and their lexicalizations. Following previous work, we use version 1.4 from \citet{ferreira2018enriching}. The E2E dataset \citep{novikova-etal-2017-e2e} contains restaurant descriptions based on sets of attributes (slots). In this work, we refer to the cleaned version of the E2E dataset \citep{dusek_semantic_2019}. For the domain adaptation experiments, we use \textsc{DiscoFuse} \citep{geva-etal-2019-discofuse}, which is a large-scale dataset for sentence fusion.

\subsection{Data Preprocessing} For WebNLG, we extract the initial templates from the training data from examples containing only a \textit{single} triple. In the E2E dataset, there are no such examples; therefore our solution is twofold: we extract the templates for \textit{pairs} of predicates, using them as a starting point for the algorithm in order to leverage the lexical variability in the data (manually filtering out the templates with semantic noise), and we also create a small set of templates for each \textit{single} predicate manually, using them in the subsequent steps of the algorithm (this is possible due to the low variability of the predicates in the dataset).\footnote{In the E2E dataset, the data is in the form of key-value slots. We transform the data to RDF triples by using the name of the restaurant as a \textit{subject} and the rest of the slots as \textit{predicate} and \textit{object}. This creates \textit{n-1} triples for \textit{n} slots.} See Table~\ref{tab:templates_ex} for examples of templates we used in our experiments.

\subsection{Setup}
As a \emph{baseline}, we generate the best templates according to \textsc{LMScorer} without applying the sentence fusion (i.e.\ always using the fallback). 

For the \emph{sentence fusion} experiments, we use \textsc{LaserTagger} with the autoregressive decoder with a beam of size 10. We use all reference lexicalizations and the vocabulary size $V=100$, following our preliminary experiments, which showed that filtering the references only by limiting the vocabulary size brings the best results (see Supplementary for details). We finetune the model for 10,000 updates with batch size 32 and learning rate $2 \times 10^{-5}$.
For the beam filtering heuristic, we check for the presence of entities by simple string matching in WebNLG; for the E2E dataset, we use a set of regular expressions from TGen\footnote{\url{https://github.com/UFAL-DSG/tgen}} \citep{dusek_semantic_2019}. We do not use any pre-ordering steps for the triples and process them in the default order.

Additionally, we conduct a \textit{zero-shot domain adaptation} experiment. We train the sentence fusion model with the same setup, but instead of the in-domain datasets, we use a subset of the balanced-Wikipedia portion of the \textsc{DiscoFuse} dataset. In particular, we use the discourse types which frequently occur in our datasets, filtering the discourse types which are not relevant for our use-case. See Supplementary for the full listing of the selected types.


\begin{table*}[ht] \small
    \begin{tabular}{l p{13.5cm}}
        \textbf{Triples}   & \textit{(Albert Jennings Fountain, deathPlace, New Mexico Territory); (Albert Jennings Fountain, birthPlace, New York City); (Albert Jennings Fountain, birthPlace, Staten Island)} \\ \midrule
       \textbf{Step \#0} & Albert Jennings Fountain died in New Mexico Territory. \\
       \textbf{Step \#1} & Albert Jennings Fountain, who died in New Mexico Territory, was born in \greenund{New York City}. \\
       \textbf{Step \#2} & Albert Jennings Fountain, who died in New Mexico Territory, was born in New York City, \greenund{Staten Island}. \\ \midrule
       \textbf{Reference} & Albert Jennings Fountain was born in Staten Island, New York City and died in the New Mexico Territory.
       \end{tabular}
    \caption{An example of correct behavior of the algorithm on the WebNLG dataset. Newly added entities are underlined, the output from Step \#2 is the output text.}\label{tab:ex0}
    \end{table*}

\section{Analysis of Results}

We compute the metrics used in the evaluation of the E2E Challenge \citep{duvsek2020evaluating}: BLEU \citep{papineni-etal-2002-bleu},  NIST \citep{doddington2002automatic}, METEOR \citep{banerjee-lavie-2005-meteor}, ROUGE$_L$ \citep{lin-2004-rouge} and CIDEr \citep{vedantam2015cider}. The results are shown in Table~\ref{tab:results}.
The scores from the automatic metrics lag behind the state-of-the-art, although both the fusion and the zero-shot approaches show improvements over the baseline. We examine the details in the following paragraphs, discussing the behavior of our approach, and we outline plans for improving the results in Section~\ref{sec:future}.

\paragraph{Accuracy vs.\ Variability} Our approach ensures zero entity errors, since the entities are filled verbatim into the templates and in case an entity is missing in the whole beam, a fallback is used instead. Semantic inconsistencies still occur, e.g.\ if a verb or function words are missing.

The fused sentences in the E2E dataset, where all the objects are related to a single subject, often lean towards compact forms, e.g.: \textit{Aromi is a family friendly chinese coffee shop with a low customer rating in riverside}. On the contrary, the sentence structure in WebNLG mostly follows the structure from the templates and the model performs minimal changes to fuse the sentences together. See Table~\ref{tab:ex0} and Supplementary for examples of the system outputs. Out of all steps, 28\% are fallbacks (no fusion is performed) in WebNLG and 54\% in the E2E dataset. 
The higher number of fallbacks in the E2E dataset can be explained by a higher lexical variability of the references, together with a higher number of data items per example in the E2E dataset, making it harder for the model to maintain the text coherency over multiple steps.

\paragraph{Templates} On average, there are 12.4 templates per predicate in WebNLG and 8.3 in the E2E dataset. In cases where the set of templates is more diverse, e.g.\ if the template for the predicate \textit{country} has to be selected from \{\textit{<subject> is situated within <object>, <subject> is a dish found in <object>}\}, \textsc{LMScorer} helps to select the semantically accurate template for the specific entities. The literal copying of entities can be too rigid in some cases, e.g.\ \textit{Atatürk Monument (İzmir) is made of ``Bronze''}, but these disfluencies can be improved in the fusion step.

\paragraph{Reordering} \textsc{LaserTagger} does not allow arbitrary reordering of words in the sentence, which can limit the expressiveness of the sentence fusion model. Consider the example in Figure~\ref{fig:alg}: in order to create a sentence \textit{English is spoken in Dublin, the capital of Ireland}, the model has to delete and re-insert at least one of the entities, e.g.\ \textit{English}, which has to be present in the vocabulary.

\paragraph{Domain Independence} The zero-shot model trained on \textsc{DiscoFuse} is able to correctly pronominalize or delete repeated entities and join the sentences with conjunctives, e.g.\ \textit{William Anders was born in British Hong Kong\underline{, and was} a member of the crew of Apollo 8}. While the model makes only a limited use of sentence fusion, it makes the output more fluent while keeping strong guarantees of the output accuracy.

\section{Future Work}
\label{sec:future}
Although the current version of our approach is not yet able to consistently produce sentences with a high degree of fluency, we believe that the approach provides a valuable starting point for controllable and domain-independent D2T generation. In this section, we outline possible directions for tackling the main drawbacks and improving the results of the model with further research.

Building a high-quality sentence fusion model, which lies at the core of our approach, remains a challenge \cite{lebanoff2020learning}. Our simple extractive approach relying on existing D2T datasets may not produce sufficient amount of clean data. On the other hand, the phenomena covered in the \textsc{DiscoFuse} dataset are too narrow for the fully general sentence fusion. We believe that training the sentence fusion model on a larger and more diverse sentence fusion dataset, built e.g. in an unsupervised fashion \cite{lebanoff-etal-2019-scoring}, is a way to improve the robustness of our approach.

Fluency of the output sentences may be also improved by allowing more flexibility for the order of entities, either by including an ordering step in the pipeline \citep{moryossef2019astep}, or by using a text-editing model that is capable of explicit reordering of words in the sentence \cite{mallinson2020felix}. Splitting the data into smaller batches (i.e. setting an upper bound for the number of sentences fused together) could also help to improve the consistency of results with a higher number of data items.

Our string matching heuristic is quite crude and may lead to a high number of fallbacks. Introducing a more precise heuristic, such as a semantic fidelity classifier \citep{harkous2020have}, or a model trained for natural language inference \citep{dusek2020nli} could help to promote lexical variability of the text.

Finally, we note that the text-editing paradigm allows to visualize the changes made by the model, introducing the option to accept or reject the changes at each step, and even build a set of custom rules on top of the individual edit operations based on the affected tokens. This flexibility could be useful for tweaking the model manually for a production system.

\section{Conclusions}
We proposed a simple and intuitive approach for D2T generation, splitting the process into two steps: lexicalization of data and improving the text fluency. A trivial lexicalization helps to promote fidelity and domain independence while delegating the subtle work with language to neural models allows to benefit from the power of general-domain pre-training. While a straightforward application of this approach on the WebNLG and E2E datasets does not produce state-of-the-art results in terms of automatic metrics, the results still show considerable improvements above the baseline. We provided insights into the behavior of the model, highlighted its potential benefits, and proposed the directions for further improvements.

\section*{Acknowledgements}
We would like to thank the anonymous reviewers for the relevant comments. The work was supported by the Charles University grant No.~140320, the SVV project No.~260575, and the Charles University project PRIMUS/19/SCI/10.

\bibliographystyle{acl_natbib}
\bibliography{acl2020}

\clearpage
\appendix

\onecolumn

\noindent{\large\bf\MyTitle: Supplementary Material}

\section{Hyperparameter Setup}
Examples in the original datasets can have multiple reference lexicalizations. We introduce three strategies for dealing with this fact during the construction of the training dataset for the sentence fusion model:

\begin{itemize}[nosep]
    \item \textit{“best”}: select the best lexicalizations for both the source and the target using \textsc{LMScorer} 
    \item \textit{“best\_tgt”}: select the best lexicalization for the target using \textsc{LMScorer} and use all lexicalizations for the source
    \item \textit{“all”}: use all lexicalizations for both the source and the target
\end{itemize}
Note that the training dataset is further filtered by the limited vocabulary of \textsc{LaserTagger}, which helps to filter out the outliers. We experiment with vocabulary sizes $V \in \{100, 500, 1000, 5000\}$. Table~\ref{tab:dev} shows the results on the development sets of both datasets. Based on these results, we select $V=100$ and the strategy \textit{all} for our final experiments.

\begin{table*}[h]
\centering\small
\begin{tabular}{l cccc cccc cccc}
& \multicolumn{12}{c}{\bf WebNLG} \\  \toprule 

& \multicolumn{4}{c}{\it best } &   \multicolumn{4}{c}{\it best\_tgt  }  &  \multicolumn{4}{c}{\it all  }        \\
\textbf{vocab. size} & 100 & 500 & 1000 & 5000 & 100 & 500 & 1000 & 5000 & 100 & 500 & 1000 & 5000  \\
\cmidrule(lr){2-5} \cmidrule(lr){6-9} \cmidrule(lr){10-13}
\textbf{BLEU} & 0.373 & 0.370 & 0.370 & 0.335 & 0.382 & 0.382 & 0.375 & 0.342 & \textbf{0.397} & 0.389 & 0.391 & 0.370 \\
\textbf{NIST} & 7.610 & 7.478 & 7.411 & 6.713 & 7.673 & 7.596 & 7.470 & 6.831 & \textbf{7.912} & 7.676 & 7.679 & 7.307 \\
\textbf{METEOR} & 0.398 & 0.399 & 0.397 & 0.396 & \textbf{0.401} & 0.399 & 0.396 & 0.393 & 0.400 & \textbf{0.401} & 0.400 & 0.399 \\
\textbf{ROUGE$_L$} & 0.566 & 0.569 & 0.568 & 0.553 & 0.569 & 0.570 & 0.569 & 0.556 & 0.574 & \textbf{0.577} & 0.576 & 0.568 \\
\textbf{CIDER} & 2.586 & 2.573 & 2.466 & 2.023 & 2.594 & 2.525 & 2.466 & 2.133 & \textbf{2.639} & 2.570 & 2.557 & 2.385 \\
\bottomrule
\\
& \multicolumn{12}{c}{\bf E2E} \\  \toprule 

& \multicolumn{4}{c}{\it best } &   \multicolumn{4}{c}{\it best\_tgt  }  &  \multicolumn{4}{c}{\it all  }        \\
\textbf{vocab. size} & 100 & 500 & 1000 & 5000 & 100 & 500 & 1000 & 5000 & 100 & 500 & 1000 & 5000  \\
\cmidrule(lr){2-5} \cmidrule(lr){6-9} \cmidrule(lr){10-13}
\textbf{BLEU} & 0.252 & 0.254 & 0.249 & 0.255 & 0.269 & 0.258 & 0.260 & 0.256 & \textbf{0.293} & 0.277 & 0.273 & 0.268 \\
\textbf{NIST} & 4.168 & 4.180 & 4.049 & 4.077 & 4.435 & 4.167 & 4.154 & 4.097 & \textbf{4.762} & 4.461 & 4.357 & 4.238 \\
\textbf{METEOR} & 0.345 & 0.346 & 0.348 & 0.351 & 0.351 & 0.352 & 0.351 & 0.350 & 0.353 & 0.350 & 0.352 & \textbf{0.355} \\
\textbf{ROUGE$_L$} & 0.426 & 0.435 & 0.429 & 0.429 & 0.441 & 0.434 & 0.435 & 0.430 & \textbf{0.460} & 0.448 & 0.447 & 0.441 \\
\textbf{CIDEr} & 0.739 & 0.759 & 0.647 & 0.634 & 0.929 & 0.728 & 0.693 & 0.678 & \textbf{1.128} & 0.967 & 0.881 & 0.799 \\
\midrule
\bottomrule
\end{tabular}
\caption{Results of automatic metrics on the WebNLG and E2E development sets with different reference strategies and vocabulary sizes.}
\label{tab:dev}
\end{table*}

\section{Discourse Types}

The list of different discourse types available in the \textsc{DiscoFuse} dataset, with an indication whether they were selected for our zero-shot training, is shown in Table~\ref{tab:df}.

\begin{table*}[h]
\centering
\begin{tabular}{lc lc}
\toprule
\textbf{type} & \textbf{selected} & \textbf{type} & \textbf{selected} \\ \midrule
\texttt{PAIR\_ANAPHORA} & yes & \texttt{SINGLE\_CONN\_INNER\_ANAPHORA} & no \\
\texttt{PAIR\_CONN} & no & \texttt{SINGLE\_CONN\_START} & no \\
\texttt{PAIR\_CONN\_ANAPHORA} & no & \texttt{SINGLE\_RELATIVE} & yes \\
\texttt{PAIR\_NONE} & yes & \texttt{SINGLE\_S\_COORD} & yes* \\
\texttt{SINGLE\_APPOSITION} & yes & \texttt{SINGLE\_S\_COORD\_ANAPHORA} & yes* \\
\texttt{SINGLE\_CATAPHORA} & no & \texttt{SINGLE\_VP\_COORD} & yes* \\
\texttt{SINGLE\_CONN\_INNER} & no & & \\
\bottomrule
\end{tabular}
\caption{A list of available discourse types in the \textsc{DiscoFuse} dataset. For our zero-shot experiments, we select a subset of \textsc{DiscoFuse}, omitting the phenomena which mostly do not occur in our datasets. The asterisk (*) symbolizes that only the examples with the connectives \textit{"and"} or \textit{", and"} were selected.}
\label{tab:df}
\end{table*}
\clearpage
\section{Output Examples}

Tables~\ref{tab:ex1}--\ref{tab:ex4} show examples of outputs of our iterative sentence fusion method (with in-domain training) on both the E2E and WebNLG datasets. We show both instances that produce flawless output (Tables~\ref{tab:ex3} and~\ref{tab:ex1}) and instances where our approach makes an error (Table~\ref{tab:ex4} and~\ref{tab:ex2}).
Table~\ref{tab:ex5} then illustrates the behavior of the zero-shot approach (without in-domain training data).

\begin{table*}[ht]
    \begin{tabular}{l p{13.5cm}}
    \toprule
        \textbf{Triples}   & \texttt{(A Loyal Character Dancer, publisher, Soho Press); (Soho Press, country, United States); (United States, leaderName, Barack Obama)} \\ 
       \textbf{Step \#0} & Soho Press is the publisher of A Loyal Character Dancer. \\
       \textbf{Step \#1} & Soho Press is the publisher of A Loyal Character Dancer which can be found in the \greenund{United States}. \\
       \textbf{Step \#2} & Soho Press is the publisher of A Loyal Character Dancer which can be found in the United States where \greenund{Barack Obama} is president. \\ 
       \textbf{Reference} & A Loyal Character Dancer is published by Soho Press in the United States where Barack Obama is the president. \\
       \bottomrule
       \end{tabular}
    \caption{An example of correct behavior of the algorithm on the WebNLG dataset (newly added entities are underlined).}\label{tab:ex3}
    \end{table*}

\begin{table*}[ht]
    \begin{tabular}{l p{13.5cm}}
        \toprule
        \textbf{Triples}   & \texttt{(Giraffe, area, riverside); (Giraffe, eatType, pub); (Giraffe, familyFriendly, no); (Giraffe, food, Chinese); (Giraffe, near, Raja Indian Cuisine}) \\
       \textbf{Step \#0} & Giraffe serves French food and is not family-friendly. \\
       & \quad \rotatebox[origin=c]{180}{$\Lsh$} \textit{A template for the pair of predicates "eatType" and "familyFriendly" is selected}. \\
       \textbf{Step \#1} & Giraffe serves French food in the \greenund{riverside} area and is not family-friendly. \\
       \textbf{Step \#2} & Giraffe is a French \greenund{pub} in the riverside area that is not family-friendly. \\
       \textbf{Step \#3} & Giraffe is a French pub in riverside that is not family-friendly. It is located near \greenund{Raja Indian Cuisine} . \\ 
       \textbf{Reference} & Giraffe is a not family-friendly French pub near Raja Indian Cuisine near the riverside. \\
       \bottomrule
       \end{tabular}
    \caption{An example of correct behavior of the algorithm on the E2E dataset (newly added entities are underlined).}\label{tab:ex1}
    \end{table*}

    \begin{table*}[ht]
        \begin{tabular}{l p{13.5cm}}
            \toprule
            \textbf{Triples}   & \texttt{(Poland, language, Polish language); (Adam Koc, nationality, Poland); (Poland, ethnicGroup, Kashubians)} \\
           \textbf{Step \#0} & Polish language is one of the languages that is spoken in Poland . \\
           \textbf{Step \#1} & Polish language is spoken in Poland, where Adam Koc \redund{is spoken}. \\
           & \quad \rotatebox[origin=c]{180}{$\Lsh$} \textit{An incorrect expression is inserted}. \\
           \textbf{Step \#2} & Polish language is spoken in Poland, where Adam Koc \redund{is spoken} and Kashubians are an ethnic group. \\
           \textbf{Reference} & The Polish language is used in Poland, where Adam koc was from. Poland has an ethnic group called Kashubians. \\
           \bottomrule
           \end{tabular}
        \caption{An example of incorrect behavior of the algorithm on the WebNLG dataset (with the error underlined).}\label{tab:ex4}
        \end{table*}

\begin{table*}[ht]
\begin{tabular}{l p{13.5cm}}
    \toprule
    \textbf{Triples}   & \texttt{(The Phoenix, area, riverside); (The Phoenix, eatType, restaurant); (The Phoenix, familyFriendly, yes); (The Phoenix, near, Raja Indian Cuisine);  (The Phoenix, priceRange, cheap)} \\ 
   \textbf{Step \#0} & The Phoenix is a cheap place to eat. Yes it is family friendly. \\
   & \quad \rotatebox[origin=c]{180}{$\Lsh$} \textit{A template for the pair of predicates "price" and "familyFriendly" is selected}. \\
   \textbf{Step \#1} & The Phoenix is a \redund{cheap family friendly on the riverside}. \\
   & \quad \rotatebox[origin=c]{180}{$\Lsh$} \textit{A grammatical error is made}. \\
   \textbf{Step \#2} & The Phoenix is a \redund{cheap family friendly offering} restaurant in the riverside area. \\
   & \quad \rotatebox[origin=c]{180}{$\Lsh$} \textit{The grammar of the sentence is still not correct}. \\
   \textbf{Step \#3} & The Phoenix is a cheap, family friendly restaurant in the riverside area, located near Raja Indian Cuisine. \\
    & \quad \rotatebox[origin=c]{180}{$\Lsh$} \textit{Grammatical errors are fixed in the last step of sentence fusion}. \\ 
   \textbf{Reference} & Cheap food and a family friendly atmosphere at The Phoenix restaurant. Situated riverside near the Raja Indian Cuisine. \\
   \bottomrule
   \end{tabular}
\caption{An example of behavior of the algorithm on the E2E dataset with several intermediate mistakes (underlined) and fixed output.}\label{tab:ex2}
\end{table*}

\begin{table*}[ht]
\begin{tabular}{l p{13.5cm}}
    \toprule
    \textbf{Triples}   & \texttt{(Arrabbiata sauce, region, Rome); (Arrabbiata sauce, country, Italy); (Arrabbiata sauce, ingredient, olive oil)} \\
   \textbf{Step \#0} & Arrabbiata sauce is a dish that comes from the Rome region. \\
   & \quad \rotatebox[origin=c]{180}{$\Lsh$} \textit{A template for the predicate "region" (suitable for food) is selected}. \\
   \textbf{Step \#1} & Arrabbiata sauce is a dish that comes from the Rome region\greenund{, and it} is a dish that is popular in Italy. \\
   & \quad \rotatebox[origin=c]{180}{$\Lsh$} \textit{The sentences are correctly joined together}. \\
   \textbf{Step \#2} & Arrabbiata sauce is a dish that comes from the Rome region, and it is a dish that is popular in Italy. Olive oil is one of the ingredients used to make Arrabbiata sauce. \\
   & \quad \rotatebox[origin=c]{180}{$\Lsh$} \textit{The text is left intact.} \\
   \textbf{Reference} & Arrabbiata sauce is a traditional dish from Rome, Italy. Olive oil is one of the ingredients in the sauce. \\
   \bottomrule
   \end{tabular}
\caption{An example of behavior of the zero-shot algorithm on the WebNLG dataset (with a single change made by the editing step underlined).}\label{tab:ex5}
\end{table*}

\end{document}